\documentclass[10pt,twocolumn,letterpaper]{article}

\usepackage{iccv}
\usepackage{times}
\usepackage{epsfig}
\usepackage{graphicx}
\usepackage{amsmath}
\usepackage{amssymb}
\usepackage{multirow}

\usepackage[pagebackref=true,breaklinks=true,letterpaper=true,colorlinks,bookmarks=false]{hyperref}

\iccvfinalcopy 

\def\iccvPaperID{9797} 

\ificcvfinal\fi

\begin{document}

\title{Detect Only What You Specify : Object Detection with Linguistic Target}

\author{Moyuru Yamada\\
Fujitsu Laboratories Ltd.\\
4-1-1 Kamikodanaka, Nakahara-ku, Kawasaki, Kanagawa 211-8588, Japan\\
{\tt\small yamada.moyuru@fujitsu.com}
}

\maketitle
\ificcvfinal\thispagestyle{empty}\fi

\begin{abstract}
Object detection is a computer vision task of predicting a set of bounding boxes and category labels for each object of interest in a given image. 
The category is related to a linguistic symbol such as 'dog' or 'person' and there should be relationships among them.
However the object detector only learns to classify the categories and does not treat them as the linguistic symbols.
Multi-modal models often use the pre-trained object detector to extract object features from the image, but the models are separated from the detector and the extracted visual features does not change with their linguistic input.
We rethink the object detection as a vision-and-language reasoning task. We then propose targeted detection task, where detection targets are given by a natural language and the goal of the task is to detect only all the target objects in a given image.
There are no detection if the target is not given.
Commonly used modern object detectors have many hand-designed components like anchor and it is difficult to fuse the textual inputs into the complex pipeline.
We thus propose Language-Targeted Detector (LTD) for the targeted detection based on a recently proposed Transformer-based detector. 
LTD is a encoder-decoder architecture and our conditional decoder allows the model to reason about the encoded image with the textual input as the linguistic context.
We evaluate detection performances of LTD on COCO object detection dataset and also show that our model improves the detection results with the textual input grounding to the visual object.
\end{abstract}

\section{Introduction}

\begin{figure}
    \centering
    \includegraphics[width=1.0\linewidth, clip]{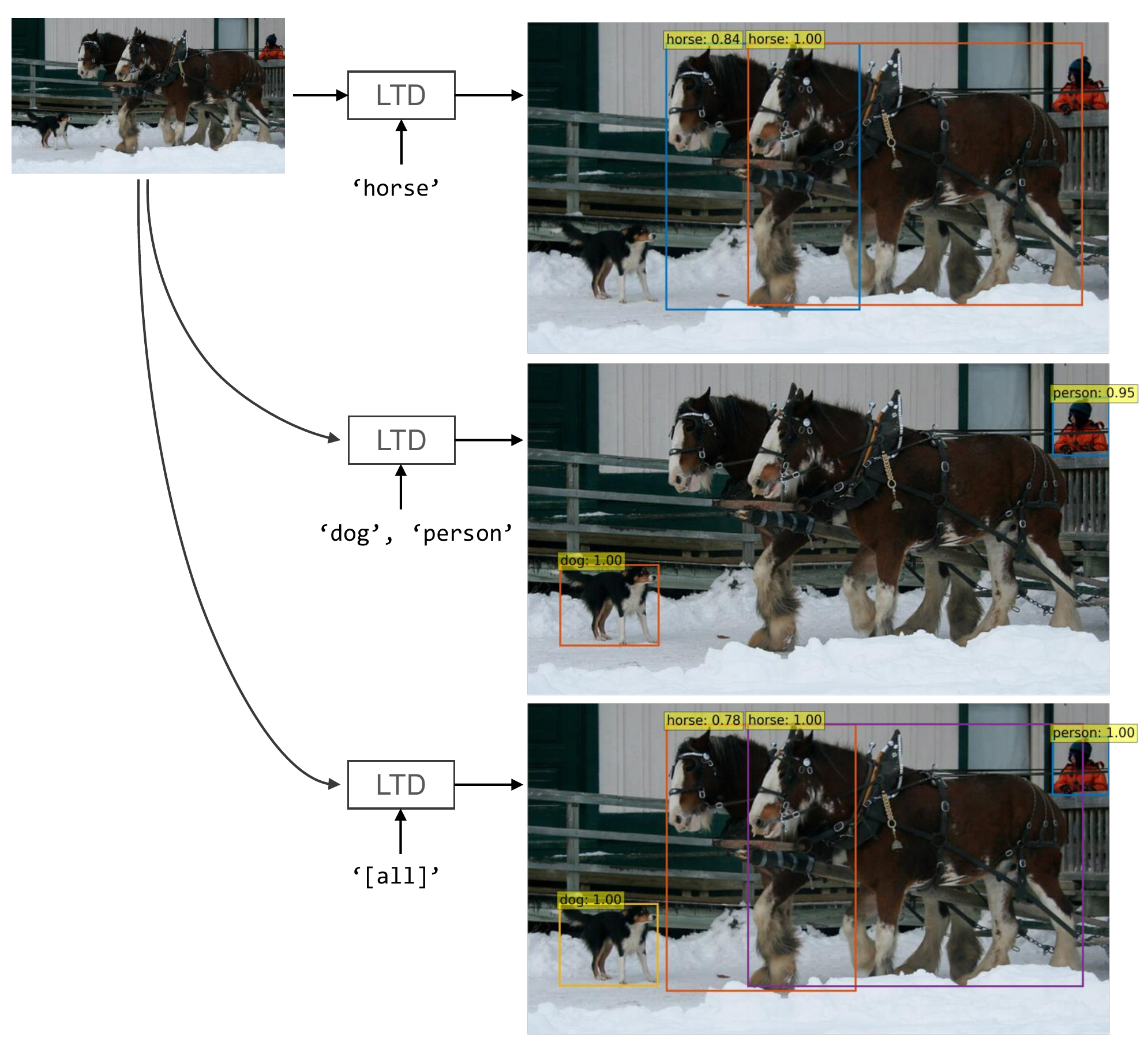}
    \caption{Language-targeted detector (LTD) detects all the target objects in an image (top). Multiple targets can be given by the natural language (middle). The detection model can also learn special tokens
    to detect specific objects, e.g., '[all]' for all objects of interest (bottom).}
    \label{figure:ltd}
\end{figure}

Object detection is a computer vision task of predicting a set of bounding boxes and category labels for each object of interest in a given image \cite{detr}. 
Each category is related to a linguistic symbol such as 'dog' or 'person', and there should be relationships among them. However the object detector only learns to classify the categories and does not treat them as the linguistic symbols. On the other hand, humans can ground the linguistic symbols to the visual objects and detect the various objects targeted by a natural language. We might also change the way of reasoning depending on what we detect, i.e., detection target. 
For example, if we are asked to count the number of dogs, we focus on dog-like shapes, or if we need to distinguish between sheep and cows, we pay attention to the differences. From the above aspect, we rethink the object detection as a vision-and-language reasoning task. As a first step, we propose targeted detection task, where detection targets are given by a natural language as a linguistic context and the goal of the task is to detect only all the target objects in a given image.

On the targeted detection, grounding the linguistic targets to the visual objects in a detection pipeline is a key challenge. Commonly used modern object detectors \cite{fast-rcnn,faster-rcnn} adopt multi-stage structure and have many hand-designed components like anchors \cite{anchor}. It is difficult to fuse the textual inputs into the complex pipeline. A recently proposed Transformer-based detector \cite{detr} has significantly simplified the pipeline. However there are no report how to design a multi-modal Transformer for the object detection.

The targeted detection is similar to referring expressions, which are natural language utterances that indicate particular objects within a scene, e.g., "the dog on the sofa" \cite{mattnet}.
Kazemzadeh \etal have proposed referring expression comprehension \cite{referitgame}, which is a task to localize the object mentioned by the given natural referring expression. However this task is typically formulated as selecting the best region from a set of proposals in an image with a textual input \cite{vilbert, mattnet}. Thus, this task does not fit for detecting multiple objects in an image, e.g., two dogs might be on the sofa. Moreover, since most works use a pre-trained object detector to generate the region proposals from the image \cite{chen2019uniter,li2019unicodervl}, the detector does not treat the textual input.

We propose Language-Targeted Detector (LTD) for the targeted detection task, as shown in Fig. \ref{figure:ltd}. A main feature of LTD is to have a textual input for specifying what objects to detect. Thus, there are no detection if the target is not given as the input. Unlike the referring expression comprehension, the targeted detection task is required to detect multiple objects with multiple targets given by the textual input. Instead of using the pre-trained object detector, we train our detection model with the textual input in end-to-end manner. We evaluate LTD on the most popular object detection dataset, COCO \cite{coco}. We show that 1) our detection model dynamically changes detected objects depending on the targets given in a natural language, and 2) by using the textual input as a context, our model improves the detection results compared to the normal object detectors. We believe the targeted detection task has the promising potential that allows the detectors to consider more complex contexts on the vision-and-language reasoning tasks.


Our contributions are as follows:
\begin{itemize}
\item We propose targeted detection task, where detection targets are given by a natural language as a linguistic context and the goal of the task is to detect only all the target objects in a given image.

\item We then propose a new object detector with a textual input for the task, and the model demonstrates that the given linguistic targets change how the detector attends to the encoded image and improve the object detection.

\end{itemize}

\begin{figure*}
    \centering
    \includegraphics[width=0.85\linewidth, clip]{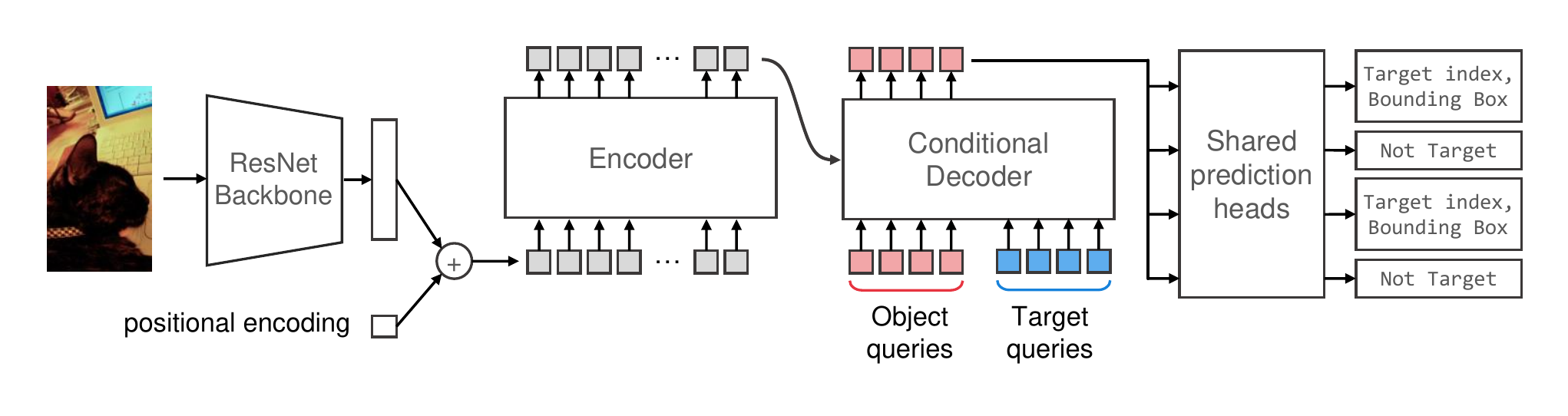}
    \caption{Overview of Language-Targeted Detector (LTD). LTD uses a CNN backbone to extract a 2D representation of an input image. It is fed into a Transformer encoder with a positional encoding. The encoder outputs the final representation of the visual features. A conditional Transformer decoder then takes a fixed number of object queries and target queries as an input sequence, and attends to the encoder outputs. The object queries are placeholders for detecting objects. The target queries are textual tokens which condition the detected objects. Shared prediction heads predict a set of bounding box, object category, and target index from an output of the decoder.
    }
    \label{figure:structure}
\end{figure*}

\section{Related Work}

\subsection{Object Detection}

After the success of deep learning on image classification \cite{dnn2,dnn1}, deep learning has been applied to object detection and many studies have continuously improved their detection performances \cite{fast-rcnn,faster-rcnn}. Deep learning-based object detectors use deep convolutional neural network (CNN) \cite{dnn2,densenet,vgg} to extract visual features from an image.
These object detectors can be categorized into two-stage, one-stage, and end-to-end ones.

Two-stage detectors such as Faster R-CNN \cite{faster-rcnn} suffer from high computational cost but achieve higher accuracy than one-stage ones, and thus their pre-trained model is often used to extract regional object features from the image for vision-and-language reasoning tasks.
Since one-stage detectors such as YOLO \cite{yolo} and SSD \cite{ssd} are faster than the two-stage ones, they fit for the real-time detection.
Both two-stage and one-stage methods need complicated and hand-designed post-processing to generate the final bounding box predictions.

Recently, Transformer-based object detection method \cite{detr} has proposed and it has significantly simplified object detection pipeline by removing the hand-crafted anchor and non-maximum suppression.
The Transformer-based detector directly predicts a set of bounding box and object category end-to-end with Hungarian bipartite matching.
It is a transformer encoder-decoder architecture and can reason about the relations of objects in the image by the attention mechanism.

These object detectors are single-modal model and take only an image as an input. They cannot alter detection targets after the training with defined object categories. Unlike them, we propose a targeted detector which enables the model to alter the detection targets by giving the targets in natural language as a linguistic context.

\subsection{Multi-Modal Transformer}
\label{subsec:multi-modal-trans}

Recent studies have shown significant improvements in natural language processing (NLP) tasks with a multi-layer Transformer \cite{transformer}, such as GPT \cite{gpt}, BERT \cite{bert}, and RoBERTa \cite{roberta}. Motivated by these promising results, the multi-layer Transformer has been utilized in multi-modal models for vision-and-language tasks. UNITER \cite{chen2019uniter} and Unicoder-VL \cite{li2019unicodervl} proposed a single-stream architecture like BERT, which fuses two modalities. ViLBERT \cite{vilbert} consists of two single-modal streams, and has a language Transfomer followed by a cross-modal Transformer. These Transformer-based multi-modal models outperform conventional state-of-the-art models \cite{dfaf,Lee_2018} on most vision and language tasks such as VQA and NLVR.

The multi-modal Transformer takes both visual features and textual tokens as input sequences. These input sequences are encoded to the image embeddings and linguistic embeddings. These embeddings are then fed into the multi-layer Transformer as an encoder. Then, they treat the above multi-modal tasks as a classification task and output answer probabilities based on the final hidden state of the Transformer encoder. Most works use a frozen pre-trained object detector to extract the regional visual features \cite{chen2019uniter,li2019unicodervl,vilbert}, and the object detectors does not consider the linguistic inputs.
Consequently, the final performances on the down-stream tasks depend on the extracted visual features. 
Therefore, instead of using outputs of the object detector, several works input grid features \cite{jiang2020defense} or pixel features \cite{pixel-bert} directly obtained from the stacked CNN layers to the Transformer encoder.
Transformer is also used for image captioning \cite{captioning}. This model is a encoder-decoder architecture and has a decoder to output a caption of a given image in addition to the encoder for the image.
We treat object detection as a vision-and-language task and propose a multi-modal decoder that conditions the detected objects by a textual input.

\subsection{Referring Expression Comprehension}

The referring expressions are natural language utterances that indicate particular objects within a scene, e.g., "the dog on the sofa" or “the book on the table”. Kazemzadeh \etal have proposed referring expression comprehension \cite{referitgame}, which is a task to localize the object mentioned by the given natural referring expression. To solve this task, some work uses CNN-LSTM structure \cite{cnn-lstm,mattnet} to encode visual and linguistic inputs. The multi-layer Transformer is also used as a multi-modal encoder and has achieved state-of-the-art result \cite{chen2019uniter,vilbert}. However these works formulate this task as selecting the best region from a set of proposals with a textual input expression and use the pre-trained object detector to generate the region proposals from a given image. Thus, the object detector is separated from the textual input. Moreover, this task does not fit for detecting multiple objects in the image.

\section{Language-Targeted Detector}

In this section, we first describe the architecture and overall pipeline of Language-Targeted Detector (LTD) and then introduce a conditional decoder.

\begin{figure*}
    \centering
    \includegraphics[width=0.8\linewidth, clip]{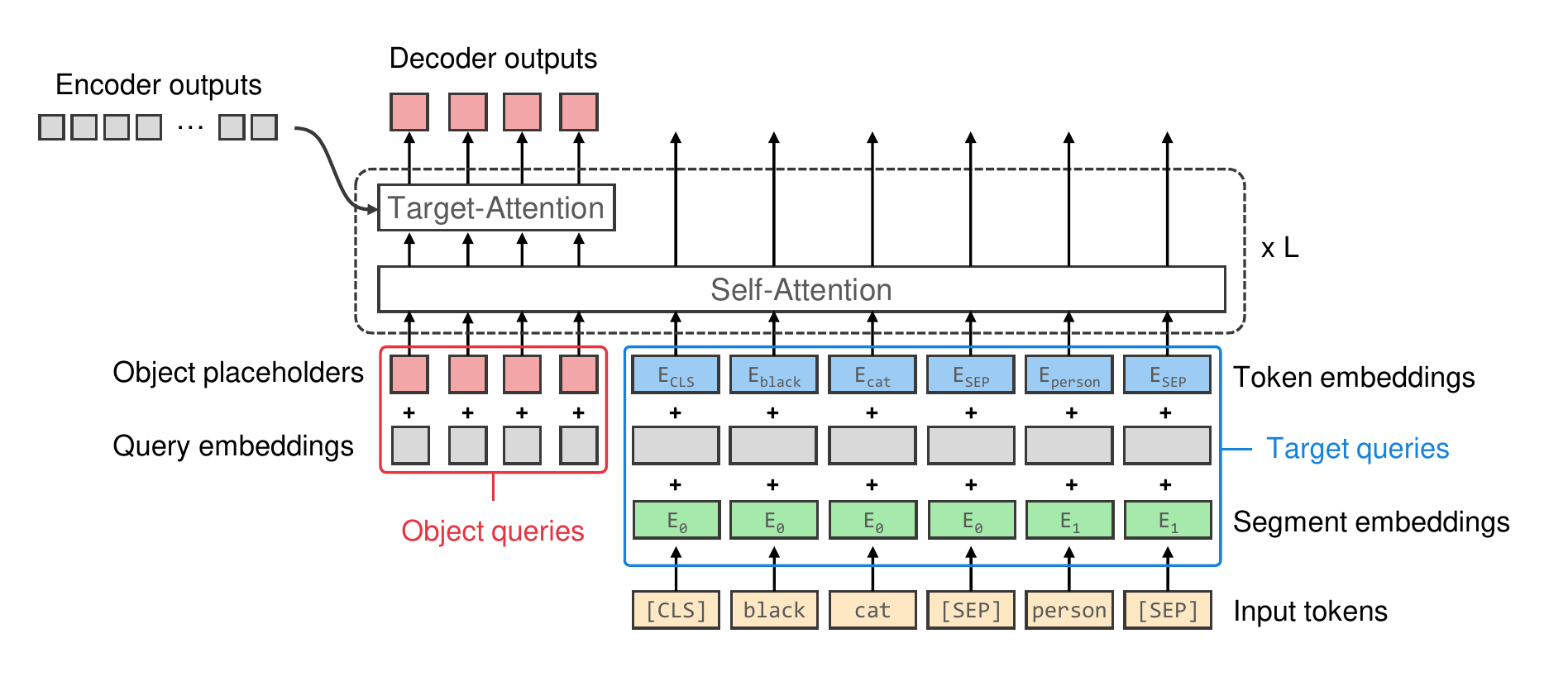}
    \caption{Conditional decoder with linguistic target. The decoder is $L$ stacks of a decoder layer composed of a self-attention layer and a target-attention layer. Each decoder layer decodes the $N$ objects with the $K$ textual tokens in parallel. The self-attention layer transforms both object queries and target queries to ground the textual tokens to the visual objects. The target-attention only treats the object queries as its attention queries. The model predicts box coordinates, class labels, and  target index using only the final output of the object queries. }
    \label{figure:decoder}
\end{figure*}

\subsection{Model Architecture}

An architecture overview  of Language-Targeted Detector (LTD) is shown in Fig. \ref{figure:structure}. 
LTD is mainly consists of a ResNet backbone, a Transformer encoder, a conditional Transformer decoder, and shared prediction heads. Most parts of the architecture inherit the Transformer-based object detector, DETR \cite{detr}. A main feature of LTD is the conditional decoder, which will be described in the section \ref{subsec:cd}.


LTD uses a Convolution Neural Network (CNN) to extract visual feature maps $f \in \mathbb{R}^{C{\times}H{\times}W}$ from an input image $I \in \mathbb{R}^{3{\times}H_0{\times}W_0}$ (3 color channels), where $H$, $W$ and $H_0$, $W_0$ are the height and width of the input image and the visual feature map, respectively. $C$ is an output channel of the visual feature map. 

The model flattens the extracted visual features and compresses the channel $C$ into a smaller dimensional representations $z \in \mathbb{R}^{d{\times}L}$ by 1$\times$1 convolution to suppress the computational cost, where $d$ is a dimension of the input visual feature $z_0$ and $L$ is its length ($L = HW$). The model collapse the spatial information of the 2D visual representation by this transformation, hence the model supplements it with a positional encoding before passing it into a Transformer encoder.


The encoder is a multi-layer Transformer. Each encoder layer has a standard Transformer block and consists of a multi-head self-attention module and a feed forward network (FFN). The encoder outputs the final representation of the visual features. 

The decoder is a stack of a self-attention layer and a target-attention layer. Each self-attention and target-attention layer are a standard Transformer architecture using multi-head attention mechanisms. Unlike DETR, our decoder takes target queries as an additional linguistic input to condition detection targets, in addition to the object queries. We design the conditional decoder to ground the target texts (e.g., 'dog') to the visual objects. The textual input is encoded to the token embeddings using a pre-trained BERT embeddings \cite{bert}.

The model have three shared prediction heads to compute the final prediction of a bounding box, a class category, and a target index. The target index indicates which object instance belongs to which target text. The bounding box is predicted by a 3-layer perceptron with ReLU activation function, whereas a linear projection layer and softmax function are used to predict the class category and the target index, respectively.

\begin{figure*}
    \centering
    \includegraphics[width=0.9\linewidth, clip]{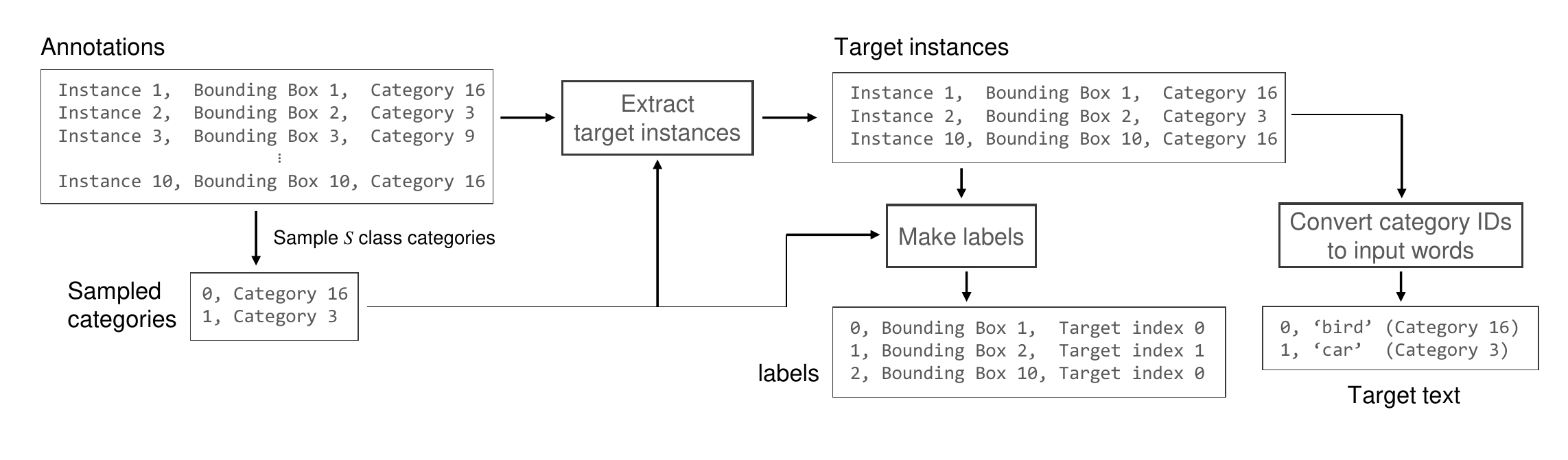}
    \caption{Data preparation procedure for the targeted detection. Target instances $\{o_1,o_2,...,o_R\}$ that belong to the randomly sampled $S$ categories are extracted from the original annotation for the image $I$. The category ID of each target object is converted to its name to obtain the target text $T$. Target index indicates which target instance belongs to which target text. We can simply choose one instance with its text annotation (i.e., class label or description) as the target instance in the image if the target dataset has no class category annotations.}
    \label{figure:learning}
\end{figure*}

\begin{figure}
    \centering
    \includegraphics[width=1.0\linewidth, clip]{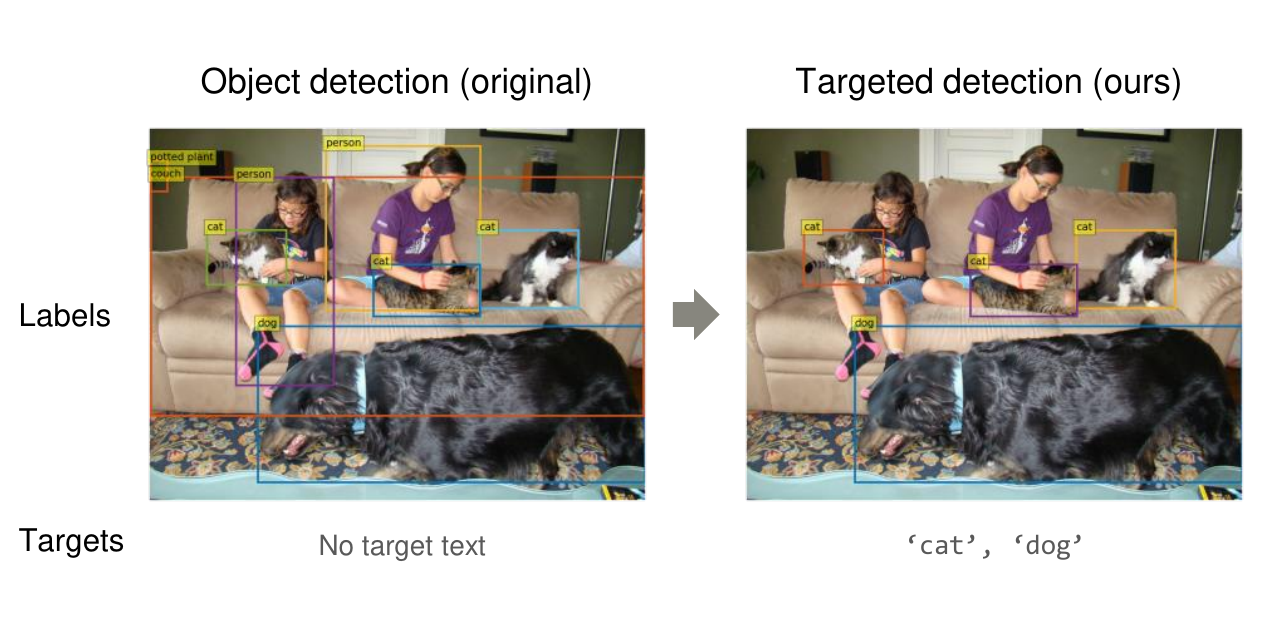}
    \caption{Examples of data conversion from the object detection to the targeted detection.
    }
    \label{figure:annotation}
\end{figure}

\subsection{Conditional Decoder}
\label{subsec:cd}

A challenge in designing language-targeted detector is how to control the objects to be detected by a linguistic input. To fuse the visual input and the linguistic input, most previous studies on the multi-modal Transformer have extended the encoder to allow both inputs \cite{li2019unicodervl,vilbert} as described in Sec \ref{subsec:multi-modal-trans}. Unlike these studies, we proposed a multi-modal decoder which conditions the detected objects by the linguistic target, as shown in Fig. \ref{figure:decoder}. This design is based on the following two reasons.

First, looking back on the Transformer-based object detector, the authors hypothesise that the encoder separates object instances in an image via global attention and simplifies object extraction and localization for the decoder, then the decoder only needs to attend to the object extremities such as heads or legs to extract the class and object boundaries \cite{detr}. We suppose that when the target object changes, the boundary of each object instance might not change, but the boundaries of interest change. In addition, the linguistic targets should be ground to the visual features at the object level not the pixel level to control the detected objects efficiently because the pixel level grounding is more difficult than the object level \cite{pixel-bert,tan-bansal-2019-lxmert}.

Then, a disadvantage of extending the encoder is the increase in a computational cost. The computational cost of the Transformer attention mechanisms is highly depend on a length of an input sequence. 
Compared with most multi-modal Transformer which treats object features with a typical maximum sequence length of 100 \cite{chen2019uniter,vilbert} as the visual inputs, the encoder of Transformer-based detector takes much longer sequence (over 1500) as the input \cite{detr}. Thus, we avoid further computational load due to the additional textual input for the encoder since the encoder is computationally intensive component.

The decoder is $L$ stacks of a decoder layer composed of a self-attention layer and a target-attention layer. Each decoder layer decodes the $N$ objects with the $K$ textual tokens in parallel. Objects are first encoded to the object queries with object placeholders and query embeddings. The object placeholder is a vector filled with zero value. Since the decoder is permutation-invariant, we add query embeddings as learnt positional encodings to produce different results for each object query. The object queries are sum of both. 
The textual input tokens always have a special head token (\verb'[CLS]'). We separate different targets with another special token (\verb'[SEP]').
The input tokens are first converted to its ID with a pre-trained tokenizer and then encoded to the target queries with token embeddings, segment embeddings, and the shared query embeddings. We use a pre-trained BERT embeddings as the token embeddings. The segment embeddings are a learned embedding indicating which target it belongs to. The query embeddings shared with the object queries are then added to obtain the target queries.

The self-attention layer treats both the object queries and the target queries to reason about the relations of objects in the image and ground the linguistic targets to the target objects. Target-attention layer extracts the visual features to correctly predict the objects. Here, the model needs to control the objects of interest depending on the linguistic target in the decoder, thus our decoder only takes the object queries as an input of the target-attention.
This decoder design contributes to suppress the deterioration of the computation speed. Some study have reported the slow convergence of DETR \cite{gao2021fast,sun2020rethinking}. From preliminary experiments, this design has also benefited this problem.

The model finally predicts box coordinates, class labels, and target indexes using only the final output of the $N$ object queries.

\section{Learning Targeted Detection}
\label{sec:learning_td}

In this section, we will describe our approach to learn the targeted detection task, where the detection targets are given by a natural language and the goal of the task is to detect only all the target objects in a given image.

\begin{table*}[th]
\begin{center}
\begin{tabular}{l|c|llllll}
\hline
Model & Detection target & AP & $AP_{50}$ & $AP_{75}$ & $AP_S$ & $AP_M$ & $AP_L$ \\ \cline{1-8}
Faster RCNN-FPN & \multirow{5}{*}{all} & 40.2 & 61.0 & 43.8 & 24.2 & 43.5 & 52.0  \\  \cline{1-1} \cline{3-8}
DETR &  & 42.0 & 62.4 & 44.2 & 20.5 & 45.8 & 61.1  \\  \cline{1-1} \cline{3-8}
LTD-IDX-PT &  & 42.4 & 62.1 & 45.2 & 21.1 & 46.0 & 60.9  \\
LTD-PT &  & 42.2 & 62.2 & 44.9 & 20.9 & 46.1 & 60.5  \\
LTD &  & 40.6 & 60.9 & 43.0 & 19.6 & 43.5 & 59.5  \\ \hline \hline
LTD-IDX-PT & \multirow{3}{*}{targeted only} & \textbf{47.4} & \textbf{70.1} & \textbf{50.1} & \textbf{25.0} & \textbf{50.4} & \textbf{66.5}  \\
LTD-PT &  & 47.1 & 69.3 & 49.9 & 24.8 & 50.1 & 66.2  \\
LTD &  & 45.8 & 69.0 & 48.1 & 24.2 & 48.4 & 66.1  \\ \cline{1-8} 
\end{tabular}
\end{center}
\caption{Performance comparison of the object detection on the COCO validation set. The top section shows comparison with Faster R-CNN and DETR with the same ResNet-50 backbone. We give the special token to our model to detect all the objects of interest in this comparison. The bottom section shows the targeted detection performances. -PT denotes the model initialized with the pre-trained weights of DETR and -IDX denotes the model trained with the target index prediction.
} 
\label{table:ap_class}
\end{table*}

\begin{table}[th]
\begin{center}
\begin{tabular}{llllll}
\hline
\multicolumn{6}{c}{LTD-IDX-PT} \\ \cline{1-6}
AP & $AP_{50}$ & $AP_{75}$ & $AP_S$ & $AP_M$ & $AP_L$ \\ \cline{1-6}
38.4 & 58.4 & 41.4 & 18.9 & 46.3 & 76.6  \\ \cline{1-6}
\end{tabular}
\end{center}
\caption{Performance of the target index prediction on the COCO validation set. The model predicts the target index of each detected object, i.e., which detected instance belongs to which target text.
} 
\label{table:ap_index}
\end{table}

To train a model for the targeted detection, we need to prepare an input image $I$ with an target text $T$ and box coordinate $B$, target index $E$ of targeted objects. To make these data for training and evaluation, we use the most popular object detection dataset, MS COCO. As showing in Fig. \ref{figure:learning}, the detailed procedure of the preparation are as follows.

1) The number of category samples $S$ $(1 \leq S \leq M)$ is randomly choose, where $M$ is the total number of categories in the image $I$.

2) We randomly sample $S$ categories from the list of categories. The categories $\{c_1,c_2,...,c_s\}$ are the targets.

3) We then extract all instances that belong to the target categories from the original annotation for the image $I$. Because there might be multiple instances that belong to the same category. These extracted instances $\{o_1,o_2,...,o_R\}$ are the target objects to be detected, where $R$ is the number of target objects ($R \geq S$).

4) To obtain the input target text $T$, the category ID of each target object is converted to its name (e.g., \verb'16' $\rightarrow$ `\verb'bird''). The obtained target texts are then tokenized as described in Sec. \ref{subsec:cd}.

5) Finally, target index is added to the labels to indicate which target instance belongs to which target text.

The class label of the each targeted object is optional. In the steps 1) - 3), we simply choose one instance with its text annotation (i.e., class label or description) as the target instance in the image if the target dataset has no class annotations. Thus, the targeted detection can be applied to not only object detection datasets such as MS COCO but also other vision-and-language datasets such as Visual Genome, which have images with regional descriptions but does not have class annotations.

Figure \ref{figure:annotation} shows a comparison of normal object detection and our targeted detection. Only $R$ instances in the images must be detected with the target text $T$ for the targeted detection task, whereas all instances of interest are the targets to be detected for the object detection task.

LTD can also learn special tokens to detect specific targets. In this paper, we use a special \verb'[all]' token to detect all objects of interest in the original annotation of the object detection (i.e., the input tokens are \verb'[[CLS],[all],[SEP]]').

\section{Experiments}

In this section, we first detail the experimental settings and then show that LTD learns to detect targeted object with the textual input as the context. We also analyze the conditional decoder of  the LTD to understand the behaviors.

\subsection{Experimental Settings}
\noindent
\textbf{Dataset.} 
We validate our proposed LTD on COCO 2017 detection dataset, which contains 118k training images and 5k validation
images, respectively. We report mAP for performance evaluation following previous research \cite{detr}. We make the target texts and the labels to be predicted from the original dataset as described in Sec. \ref{sec:learning_td}. When there is no annotated object in a given image, the target text is none and the input tokens are set to \verb'[[CLS],[SEP]]'. To compare the detection performances with the normal object detectors, we set a special \verb'[all]' token as the target text and simply use the all objects of interest as the targets in the image with a probability of 50\%.

\noindent
\textbf{Implementation details.}
We follow the experimental settings in the original DETR \cite{detr}. The number of the encoder layers and the decoder layers is 6 each. We use a pre-trained ResNet-50 as our backbone and an extracted 2048 dimensional visual feature is compressed to 256 dimensional input to the encoder. Our evaluation model have 100 object queries and 40 target queries. We use a pre-trained BERT-BASE tokenizer and embeddings to obtain the token embeddings from the target text. The final output of each object query is used for predicting the bounding box, the class category, and the target index. The model is trained by minimizing a L1 loss of the bounding box and a cross-entropy classification loss of the class category and the target index. The learning rate is set as $10^{-4}$ for the Transformer encoder-decoder and $10^{-5}$ for the pre-trained ResNet backbone and optimized by AdamW optimizer. All models are trained for 300 epochs on 32 V100 GPUs with 16GB memory within 3 days.

\begin{figure*}
    \centering
    \includegraphics[width=1.0\linewidth, clip]{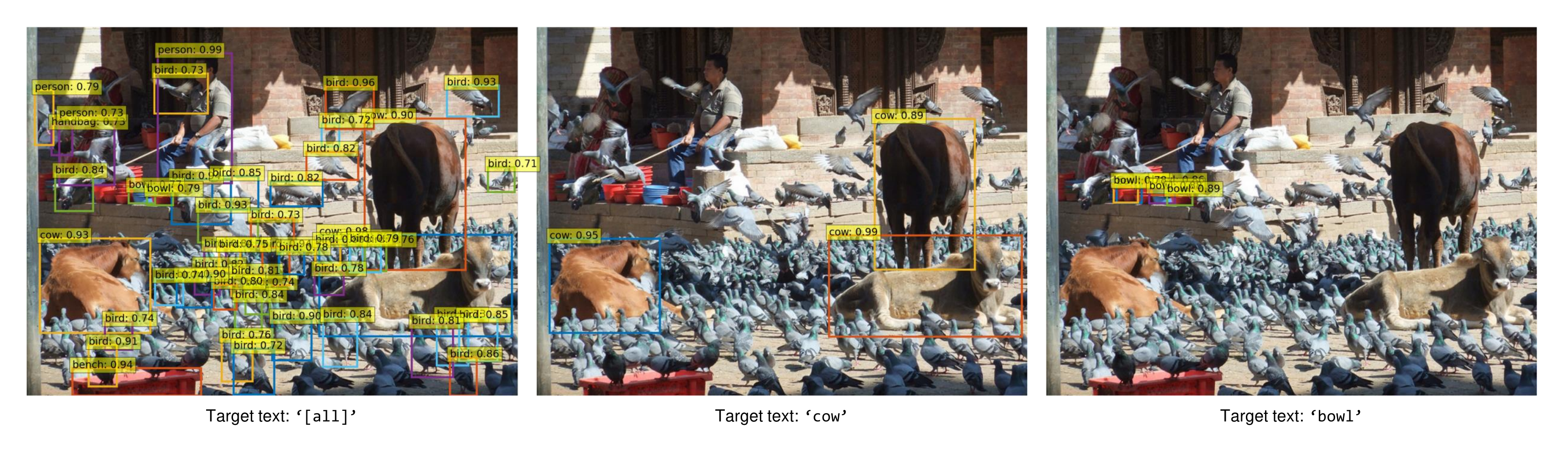}
    \caption{LTD dynamically changes the detected objects depending on the target text. LTD can detect all objects with a special token (left) and a specific target (center). Unlike just filtering the detected objects, our model detects the objects in consideration of the linguistic context (left vs right). With the target text of 'bowl', LTD detects more bowls because the model targets them.
    } 
    \label{figure:examples}
\end{figure*}

\begin{figure*}
    \centering
    \includegraphics[width=1.0\linewidth, clip]{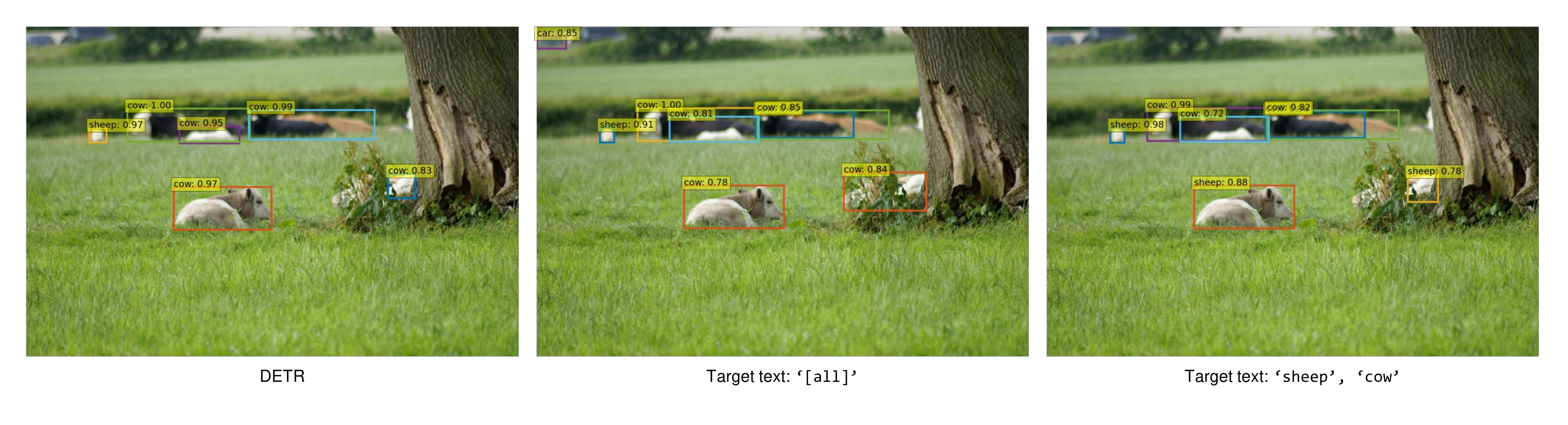}
    \caption{DETR misclassifies the objects in the foreground (left, two sheep are misclassified as cows). LTD also outputs the same incorrect results when we give the special token to detect all objects (center). On the other hand, LTD can also use the target text as the context (i.e., the targets are sheep or cows) to lead the correct answer (right).
    }
    \label{figure:examples_compare}
\end{figure*}

\subsection{Detection Performances}

We first compare the accuracy with Faster R-CNN and DETR to investigate the effect of the textual input.
Table \ref{table:ap_class} shows the results, where all models use the same ResNet-50 backbone. -PT denotes the model initialized with the pre-trained weights of DETR. -IDX denotes the model trained with the target index classification. We give the special token \verb'[all]' to detect all the objects of interest in this comparison (denoted as 'all' in the Detection target column). The results show that our models achieve comparable performances to the normal object detection models and no significant performance drop by the textual input. The initialization with the pre-trained weights of DETR is benefit for our model to speed up the training.

The bottom section in Table \ref{table:ap_class} shows the accuracy on the targeted detection task. The performances on this task are significantly better than the ones on the normal object detection on all metrics. Note that LTD learns to detects all objects of interest with the special token and does not simply refer the target text to classify the objects. Table \ref{table:ap_index} shows the accuracy of the target index prediction on the targeted detection. The model predicts which detected instance belongs to which target text. This prediction seems to be more difficult than the category classification because grounding the target text to the visual object is required.

As shown in Fig. \ref{figure:examples}, LTD dynamically changes the detected objects depending on the target text. LTD can detect all objects with the special token (left) and specific targets (center). Unlike just filtering the detected objects, our model detects the objects in consideration of the linguistic context (left vs right). With the target text of 'bowl', LTD detects more bowls because the model targets them. Figure \ref{figure:examples_compare} shows that our detection model improves the detection results using the target text as the linguistic context. DETR misclassifies the objects in the foreground (left, two sheep are misclassified as cows). LTD also outputs the same incorrect results when we give the special token to detect all objects (center). On the other hand, when we give the target text as the context (i.e., the targets are sheep or cows), our model can use the context to lead the correct answer (right).

\subsection{Analysis of Conditional Decoder}

\begin{figure*}
    \centering
    \includegraphics[width=1.0\linewidth, clip]{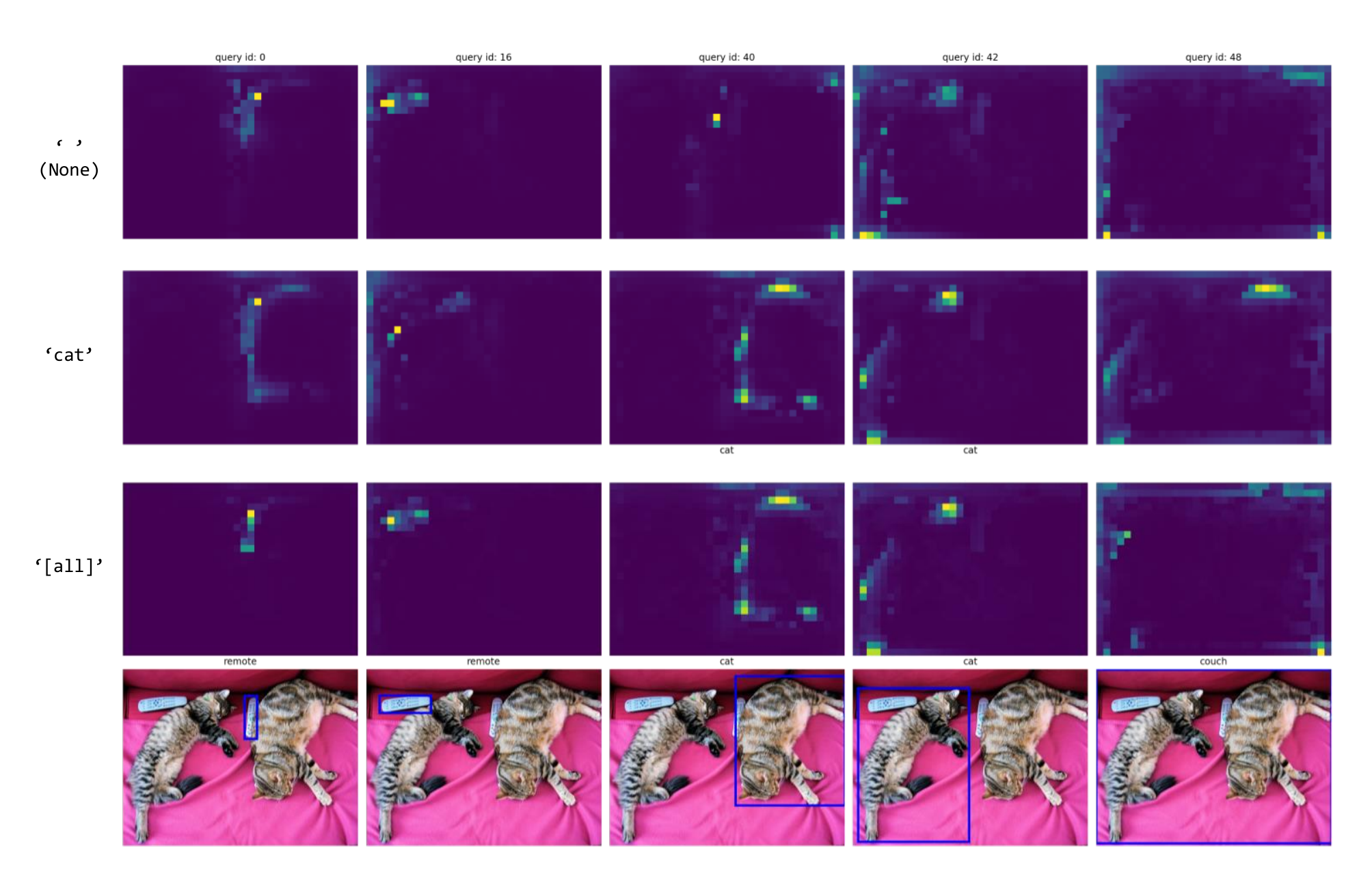}
    \caption{Each column represents a object query (queries 0, 16, 40, 42, 48th). With a special input token '[all]', LTD detects five objects, which correspond to the bottom five pictures and the five images above them that represent the predicted bounding box and the attention map of each object query. The text below each image denotes the predicted class category and the objects are not detected when there is no text there. When we give the target text 'cat', the decoder seems to attend to cat-like shapes.}
    \label{figure:analysis_tgt}
\end{figure*}

\begin{figure*}
    \centering
    \includegraphics[width=1.0\linewidth, clip]{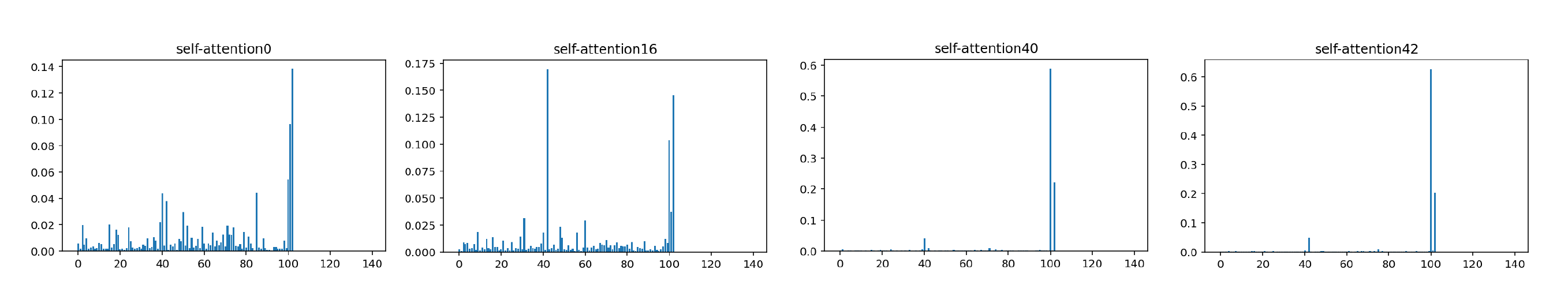}
    \caption{Self-attention weights of the object queries (0, 16, 40, 42th) when we give the target text 'cat'. The 40th and 42th queries strongly attend to the target queries ($\geq$ 100th) and they are detected as the cat.}
    \label{figure:analysis_self}
\end{figure*}

The main feature of LTD is the conditional decoder described in \ref{subsec:cd}. To understand its behavior, we visualize the target-attention map of the last decoder layer in Fig. \ref{figure:analysis_tgt} and the self-attention map of the middle decoder layer in Fig. \ref{figure:analysis_self}, respectively.

In Fig. \ref{figure:analysis_tgt}, each column represents an object query extracted from a total of 100 queries (here queries 0, 16, 40, 42, 48th). Each object query learns to specialize on certain areas and box sizes \cite{detr}. In this example, with a special input token \verb'[all]' to detect all objects, LTD detects five objects, which correspond to the bottom five pictures and the five images above them that represent the predicted bounding box and the attention map of each object query. The text below each image denotes the predicted class category. The objects are not detected when there is no text there. The attention maps show that each query mostly attends to the object extremities and boundaries. Our decoder has the linguistic target and thus the attention maps change depend on the target text. When we give the target text \verb'cat', the decoder seems to attend to cat-like shapes because the attention map of 0th and 16th queries looks similar to those of 16th and 42th queries which are classified as a cat.

Figure \ref{figure:analysis_self} shows the attention weights between the object queries and target queries when we give the target text \verb'cat'. Each image correspond to the attention weight of each query (0, 16, 40, 42th). The 40th and 42th queries strongly attend to the target queries ($\geq$ 100th) and they are detected as the cat. We hypothesize that by using the linguistic targets grounding to the objects, each object query can transform itself into a representation suitable for detecting the target, and also they can check whether they are the target.

\section{Conclusions}
We introduced the linguistic contexts into the object detection and proposed targeted detection task, where detection targets are given by natural language and the goal of the task is to detect only all the target objects in an image. We also proposed Language-Targeted Detector (LTD) for the proposed task based on the recently proposed Transformer-based detector. 
We evaluate LTD on COCO object detection dataset and show that our model improves the detection performances using the target text grounding to the visual objects as the linguistic context. 
Unlike the conventional uni-modal detectors, the conditional decoder of LTD allows the model to reason about the encoded image with the linguistic targets and dynamically control the detected objects. We expect the object detector with the linguistic context is a new direction for the vision-and-language reasoning tasks.


{\small
\bibliographystyle{ieee_fullname}
\bibliography{egbib}

\begin{thebibliography}{10}\itemsep=-1pt

\bibitem{detr}
Nicolas Carion, Francisco Massa, Gabriel Synnaeve, Nicolas Usunier, Alexander
  Kirillov, and Sergey Zagoruyko.
\newblock End-to-end object detection with transformers.
\newblock In Andrea Vedaldi, Horst Bischof, Thomas Brox, and Jan{-}Michael
  Frahm, editors, {\em Computer Vision - {ECCV} 2020 - 16th European
  Conference, Glasgow, UK, August 23-28, 2020, Proceedings, Part {I}}, volume
  12346 of {\em Lecture Notes in Computer Science}, pages 213--229. Springer,
  2020.

\bibitem{coco}
Xinlei Chen, Tsung-Yi~Lin Hao~Fang, Ramakrishna Vedantam, Saurabh Gupta, Piotr
  Dollár, and C.~Lawrence Zitnick.
\newblock Microsoft coco captions: Data collection and evaluation server, 2015.
\newblock \textit{arXiv preprint arXiv:1504.00325}.

\bibitem{chen2019uniter}
Yen-Chun Chen, Linjie Li, Licheng Yu, Ahmed~El Kholy, Faisal Ahmed, Zhe Gan, Yu
  Cheng, and Jingjing Liu.
\newblock Uniter: Universal image-text representation learning, 2019.
\newblock \textit{arXiv preprint arXiv:1909.11740}.

\bibitem{bert}
Jacob Devlin, Ming-Wei Chang, Kenton Lee, and Kristina Toutanova.
\newblock {BERT}: Pre-training of deep bidirectional transformers for language
  understanding.
\newblock In {\em Proceedings of the Conference of the North {A}merican Chapter
  of the Association for Computational Linguistics: Human Language
  Technologies}, pages 4171--4186, 2019.

\bibitem{dfaf}
Peng Gao, Zhengkai Jiang, Haoxuan You, Pan Lu, Steven C.~H. Hoi, Xiaogang Wang,
  and Hongsheng Li.
\newblock Dynamic fusion with intra- and inter-modality attention flow for
  visual question answering.
\newblock In {\em Proceedings of the IEEE Conference on Computer Vision and
  Pattern Recognition (CVPR)}, 2019.

\bibitem{gao2021fast}
Peng Gao, Minghang Zheng, Xiaogang Wang, Jifeng Dai, and Hongsheng Li.
\newblock Fast convergence of detr with spatially modulated co-attention.
\newblock 2021.
\newblock \textit{arXiv preprint arXiv:2101.07448}.

\bibitem{fast-rcnn}
R. {Girshick}.
\newblock Fast r-cnn.
\newblock In {\em IEEE International Conference on Computer Vision (ICCV)},
  pages 1440--1448, 2015.

\bibitem{dnn2}
K. He, X. Zhang, S. Ren, and J. Sun.
\newblock Deep residual learning for image recognition.
\newblock In {\em 2016 IEEE Conference on Computer Vision and Pattern
  Recognition (CVPR)}, pages 770--778, 2016.

\bibitem{captioning}
Simao Herdade, Armin Kappeler, Kofi Boakye, and Joao Soares.
\newblock Image captioning: Transforming objects into words.
\newblock In H. Wallach, H. Larochelle, A. Beygelzimer, F. d\textquotesingle
  Alch\'{e}-Buc, E. Fox, and R. Garnett, editors, {\em Advances in Neural
  Information Processing Systems}, volume~32, 2019.

\bibitem{densenet}
Gao Huang, Zhuang Liu, Laurens Van Der~Maaten, and Kilian~Q. Weinberger.
\newblock Densely connected convolutional networks.
\newblock In {\em Proceedings of the IEEE Conference on Computer Vision and
  Pattern Recognition (CVPR)}, 2017.

\bibitem{pixel-bert}
Zhicheng Huang, Zhaoyang Zeng, Bei Liu, Dongmei Fu, and Jianlong Fu.
\newblock Pixel-bert: Aligning image pixels with text by deep multi-modal
  transformers.
\newblock {\em CoRR}, abs/2004.00849, 2020.

\bibitem{jiang2020defense}
Huaizu Jiang, Ishan Misra, Marcus Rohrbach, Erik Learned-Miller, and Xinlei
  Chen.
\newblock In defense of grid features for visual question answering.
\newblock In {\em IEEE Conference on Computer Vision and Pattern Recognition
  (CVPR)}, 2020.

\bibitem{referitgame}
Sahar Kazemzadeh, Vicente Ordonez, Mark Matten, and Tamara Berg.
\newblock {R}efer{I}t{G}ame: Referring to objects in photographs of natural
  scenes.
\newblock In {\em Proceedings of the 2014 Conference on Empirical Methods in
  Natural Language Processing ({EMNLP})}, pages 787--798, Doha, Qatar, Oct.
  2014. Association for Computational Linguistics.

\bibitem{dnn1}
Alex Krizhevsky, Ilya Sutskever, and Geoffrey~E. Hinton.
\newblock Imagenet classification with deep convolutional neural networks.
\newblock {\em Commun. ACM}, 60(6):84–90, 2017.

\bibitem{Lee_2018}
Kuang-Huei Lee, Xi Chen, Gang Hua, Houdong Hu, and Xiaodong He.
\newblock Stacked cross attention for image-text matching.
\newblock In {\em Proceedings of the European Conference on Computer Vision
  (ECCV)}, 2018.

\bibitem{li2019unicodervl}
Gen Li, Nan Duan, Yuejian Fang, Ming Gong, Daxin Jiang, and Ming Zhou.
\newblock Unicoder-vl: A universal encoder for vision and language by
  cross-modal pre-training.
\newblock In {\em Proceedings of the AAAI Conference on Artificial Intelligence
  (AAAI)}, 2020.

\bibitem{anchor}
Tsung{-}Yi Lin, Priya Goyal, Ross~B. Girshick, Kaiming He, and Piotr
  Doll{\'{a}}r.
\newblock Focal loss for dense object detection.
\newblock In {\em IEEE International Conference on Computer Vision (ICCV)},
  pages 2999--3007, 2017.

\bibitem{ssd}
Wei Liu, Dragomir Anguelov, Dumitru Erhan, Christian Szegedy, Scott~E. Reed,
  Cheng{-}Yang Fu, and Alexander~C. Berg.
\newblock {SSD:} single shot multibox detector.
\newblock In {\em Proceedings of the European Conference on Computer Vision
  (ECCV)}, 2016.

\bibitem{roberta}
Yinhan Liu, Myle Ott, Naman Goyal, Jingfei Du, Mandar Joshi, Danqi Chen, Omer
  Levy, Mike Lewis, Luke Zettlemoyer, and Veselin Stoyanov.
\newblock Roberta: A robustly optimized bert pretraining approach, 2019.
\newblock \textit{arXiv preprint arXiv:1907.11692}.

\bibitem{vilbert}
Jiasen Lu, Dhruv Batra, Devi Parikh, and Stefan Lee.
\newblock Vilbert: Pretraining task-agnostic visiolinguistic representations
  for vision-and-language tasks.
\newblock In {\em Advances in Neural Information Processing Systems 32}, pages
  13--23. 2019.

\bibitem{cnn-lstm}
Ruotian Luo and Gregory Shakhnarovich.
\newblock Comprehension-guided referring expressions.
\newblock In {\em Proceedings of the IEEE Conference on Computer Vision and
  Pattern Recognition (CVPR)}, July 2017.

\bibitem{gpt}
Alec Radford, Karthik Narasimhan, Tim Salimans, and Ilya Sutskever.
\newblock Improving language understanding by generative pre-training.
\newblock 2018.

\bibitem{yolo}
Joseph Redmon, Santosh Divvala, Ross Girshick, and Ali Farhadi.
\newblock You only look once: Unified, real-time object detection.
\newblock In {\em Proceedings of the IEEE Conference on Computer Vision and
  Pattern Recognition (CVPR)}, 2016.

\bibitem{faster-rcnn}
Shaoqing Ren, Kaiming He, Ross Girshick, and Jian Sun.
\newblock Faster r-cnn: Towards real-time object detection with region proposal
  networks.
\newblock {\em IEEE Transactions on Pattern Analysis and Machine Intelligence
  (PAMI)}, 39, 2015.

\bibitem{vgg}
Karen Simonyan and Andrew Zisserman.
\newblock Very deep convolutional networks for large-scale image recognition.
\newblock In {\em Proceedings of the International Conference on Learning
  Representations (ICLR)}, 2015.

\bibitem{sun2020rethinking}
Zhiqing Sun, Shengcao Cao, Yiming Yang, and Kris Kitani.
\newblock Rethinking transformer-based set prediction for object detection.
\newblock 2020.
\newblock \textit{arXiv preprint arXiv:2011.10881}.

\bibitem{tan-bansal-2019-lxmert}
Hao Tan and Mohit Bansal.
\newblock {LXMERT}: Learning cross-modality encoder representations from
  transformers.
\newblock In {\em Proceedings of the 2019 Conference on Empirical Methods in
  Natural Language Processing and the 9th International Joint Conference on
  Natural Language Processing (EMNLP-IJCNLP)}, pages 5100--5111, 2019.

\bibitem{transformer}
Ashish Vaswani, Noam Shazeer, Niki Parmar, Jakob Uszkoreit, Llion Jones,
  Aidan~N Gomez, \L~ukasz Kaiser, and Illia Polosukhin.
\newblock Attention is all you need.
\newblock In {\em Advances in Neural Information Processing Systems 30}, pages
  5998--6008. 2017.

\bibitem{mattnet}
Licheng Yu, Zhe Lin, Xiaohui Shen, Jimei Yang, Xin Lu, Mohit Bansal, and
  Tamara~L. Berg.
\newblock Mattnet: Modular attention network for referring expression
  comprehension.
\newblock In {\em Proceedings of the IEEE Conference on Computer Vision and
  Pattern Recognition (CVPR)}, June 2018.

\bibitem{zhang2021vinvl}
Pengchuan Zhang, Xiujun Li, Xiaowei Hu, Jianwei Yang, Lei Zhang, Lijuan Wang,
  Yejin Choi, and Jianfeng Gao.
\newblock Vinvl: Revisiting visual representations in vision-language models.
\newblock In {\em CVPR 2021}, 2021.

\end{thebibliography}
}

\clearpage
\appendix
\section*{\LARGE Appendix}

\section{Dataset Statistics}

\begin{figure*}[!t]
    \centering
    \includegraphics[width=1.0\linewidth, clip]{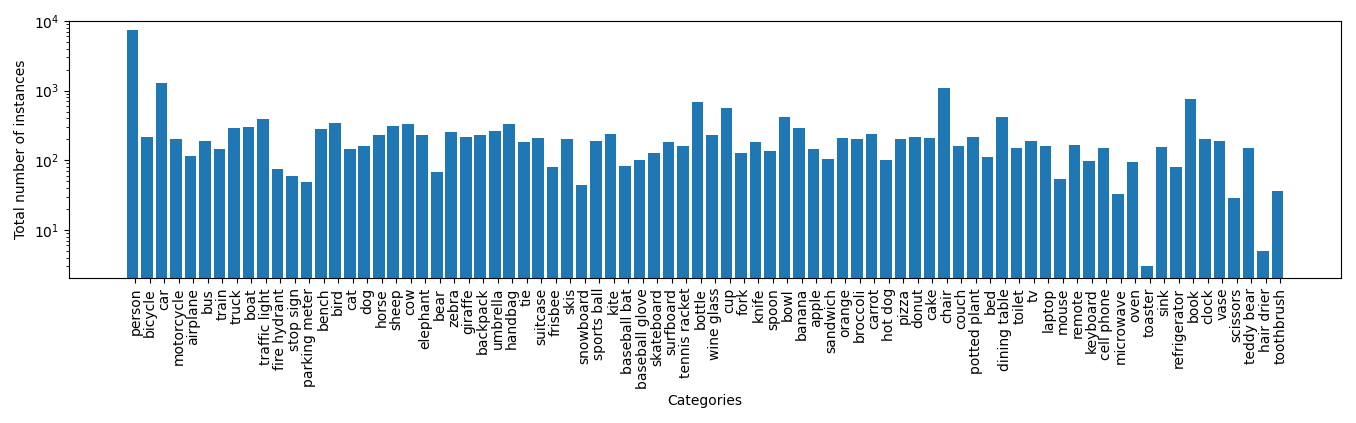}
    \caption{Total number of instances per category.}
    \label{figure:ins-cat}
\end{figure*}

\begin{figure*}[!t]
    \centering
    \includegraphics[width=1.0\linewidth, clip]{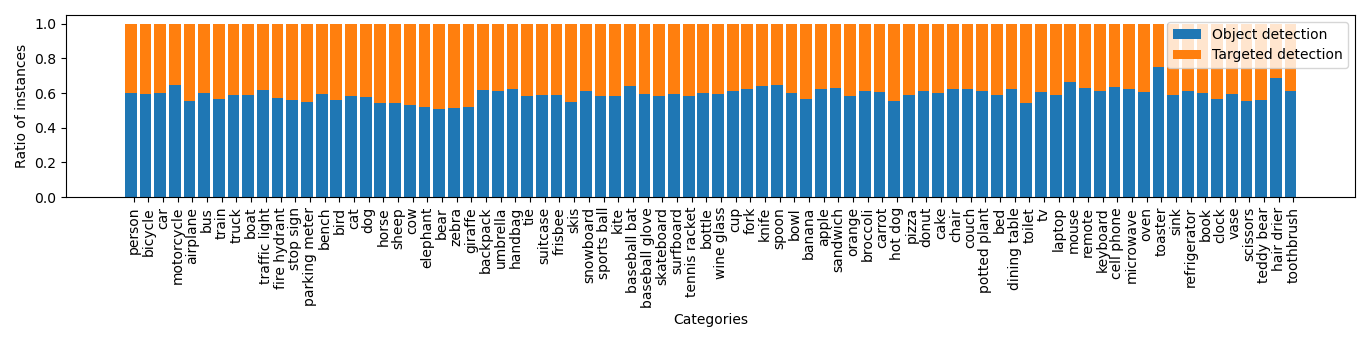}
    \caption{Ratio of instances per category of object detection and targeted detection.}
    \label{figure:ratio-ins-cat}
\end{figure*}

\begin{figure*}[t]
    \centering
    \includegraphics[width=0.8\linewidth, clip]{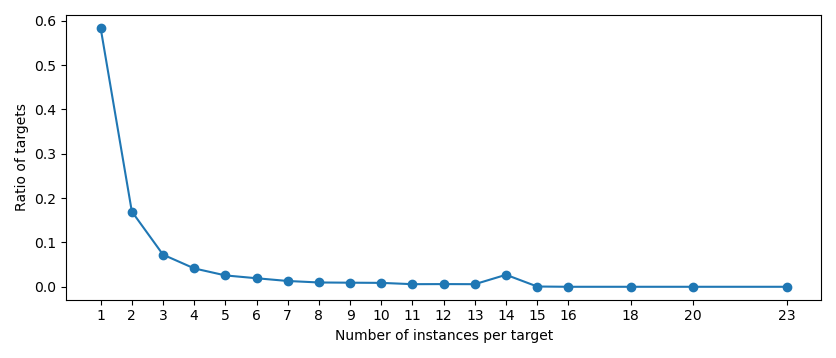}
    \caption{Ratio of the number of instances per target on targeted detection.}
    \label{figure:ratio-ins-tgt}
\end{figure*}

\begin{table}[ht]
\begin{center}
\begin{tabular}{l|c|c}
\hline
\multirow{2}{*}{} & Object & Targeted \\
 & detection &  detection \\ \cline{1-3}
Validation images & 5,000 & 5,000 \\ \cline{1-3}
Total annotated instances & 36,781 & 24,790 \\ \cline{1-3}
Categories per image & 2.93 & 1.97 \\ \cline{1-3}
Instances per image & 7.36 & 4.96  \\ \cline{1-3}
Instances per category & 2.51 & 2.51  \\ \cline{1-3}
\end{tabular}
\end{center}
\caption{Main properties of COCO object detection dataset and our targeted detection for validation. We randomly sample the object instances from the original annotation on the object detection, thus the number of instances and categories are slightly decreased.
} 
\label{table:stats}
\end{table}

We show the properties of the validation data for the target detection task. The training and evaluation data are both based on COCO object detection dataset \cite{coco}. The data preparation procedure is described in Sec. \ref{sec:learning_td} in the main paper. 

Table \ref{table:stats} shows comparisons of the main properties of COCO object detection dataset and our targeted detection. Both are validation data. Original COCO object detection dataset has 36,781 annotated instances on the validation data, whereas our targeted detection data has 24,790 instances. We randomly sample the object instances from the original annotation on the object detection, thus the number of instances and categories are slightly decreased (about 33\%). Both data contains the same 5,000 images for the validation. On average COCO dataset contains 2.93 categories and 7.36 instances per image. Our targeted detection data contains 1.97 categories and 4.96 instances per image. The number of instances per category are almost same for both data (2.51).

Figure \ref{figure:ins-cat} shows the total number of the instances per category and Fig. \ref{figure:ratio-ins-cat} shows the ratio of the instances per category of object detection and targeted detection. These results mean that the distribution of the instances in each category does not change from the original COCO dataset and the number of 'person' instances is very large compared to others.

Figure \ref{figure:ratio-ins-tgt} shows ratio of the number of instances per target on our targeted detection data. Most images contains 5 or less instances for each target and about 58\% of the targets are 1 instance. Only one image contains 23 instances for a specific target.


\section{Object Detection Conditioned on Context}
A goal of the language-targeted detector is to improve the object detection accuracy with a linguistic context, like an object label, a referring expression, or a question. 
Current SOTA multi-modal models typically use a pre-trained object detector and their performances heavily rely on the detector \cite{zhang2021vinvl}.
The detector is trained to avoid misdetection but whether the confusing object (e.g., small or occluded object) should be detected depends on the context. For instance, if the model is asked about a dog, the detector should extract the dog-like objects even if they are blurry.
However the traditional detector extracts regions of interest from an image regardless of the given linguistic context and this makes it challenging to detect the confusing object. 

As a first step, we have proposed a new object detector that detects objects conditioned on target labels and investigated how the linguistic context helps the object detection. 

\section{Comparison with Traditional Detector}

Although a main advantage of our language-targeted detector (LTD) is to detect objects in an image conditioned on target texts, in 'all' setting in Table 1, we give a constant target text '[all]' regardless of the input image to verify the baseline accuracy of LTD. This setting means our model is not conditioned on the target text, thus the performances in this setting must be comparable with DETR. \cite{gao2021fast} reports that the DETR suffers from slow convergence. Our LTD are based on the DETR and needs longer training epoch than the original DETR. Thus, we use a pre-trained weights of DETR. Without it (LTD in Table 1), the performances slightly drop due to lack of the training.

LTD detects the objects conditioned on target texts. 
In 'targeted only' setting in Table 1, LTD uses the given target text to change how the detector attends to the encoded image (Fig. 8) and thus LTD improves the object detection accuracy. Unlike just filtering regions output by the traditional detector, with the target text (e.g., 'bowl'), LTD can detect small objects that are missed without conditioning (Fig. 6).
Note that the object instances in the image are randomly sampled (line 503-520) and the category distribution is the same as one of the original dataset (Fig. 11 in Appendix). We do not choose the easy instances.

In the training, we randomly set '[all]' token as  the target text to detect all objects of interest in the image with a probability of 50\% (see 5.1). Thus, LTD does not just use the target text as a hint to predict a class category.
In Fig. 7, without conditioning (i.e., target text is '[all]'), LTD misclassifies the sheep as cow. When the target texts are 'sheep' and 'cow', although these cannot be a direct hints, LTD uses the textual input and correctly predicts the object categories (see also Fig. 9).

Thank you for suggesting several tests to verify our LTD. We have already conducted some of them. We would like to add those results in the final paper. First, in 'targeted only' setting, we add deceptive targets that does not appear in the image.  The number of the deceptive targets are set to $max(1, N*r)$ where $N$ is the number of the ground-truth targets and $r$ is rate. The accuracy dropped by around 1 point with 10\% rate. There is no further drop at 20\%. 
Then, in 'all' setting, we give all ground-truth targets that appear in each image as the context. The accuracy increases by around 5 points and reaches the same as performances in 'targeted only' setting because of the same data distribution.
\section{Experimental Settings and Others}

\subsection{Target texts}
Target text such as 'dog' or 'bird' is sampled from the ground-truth annotations (see Fig.4). We described the detection performances with the deceptive target that does not appears in the ground-truth above.

\subsection{Training objectives}
Prediction of the class category is optional but we use it during the training of LTD because it improve the convergence speed. We use the class category in COCO for performance comparisons. The total loss can be written as follows:
\begin{equation}
    L = k_{C}L_{C} + k_{B}L_{B} + k_{I}L_{I}
\end{equation}
where $L_{n}, k_{n} (n={C, B, I})$ are a loss and a coefficient of the class category, the bounding box, and the target index.

\subsection{Generalization of model}
To easily extend LTD for referring expression comprehension on Flickr30K Entities and Visual Genome, we use BERT to encode the textual input to token embeddings (see 3.2). LTD can take not only the object labels but also longer text like sentences as an input.

\end{document}